\documentclass[10pt, a4paper]{article}

\usepackage{lrec-coling2024}

\usepackage{multirow} 
\usepackage{tabularx} 
\usepackage{adjustbox} 
\usepackage{booktabs} 
\usepackage{makecell} 

\usepackage{tcolorbox} 
\usepackage{float} 

\usepackage{blindtext}
\usepackage{hyperref}
\usepackage{amsmath}

\title{PSYDIAL: Personality-based Synthetic Dialogue Generation using Large Language Models\\}

\name{Ji-Eun Han\(^{1,2}\), Jun-Seok Koh\(^{1}\), Hyeon-Tae Seo\(^{1}\), Du-Seong Chang\(^{1}\), Kyung-Ah Sohn\(^{2,*}\)} 
\address{
        \(^{1}\)KT \\ \(^{2}\)Department of Artificial Intelligence, Ajou University \\
         \{ji-eun.han, js.koh, ht.seo,  dschang\}@kt.com\\
         \{kasohn\}@ajou.ac.kr\\
         \text{* Corresponding author}\\}

\abstract{
We present a novel end-to-end personality-based synthetic dialogue data generation pipeline, specifically designed to elicit responses from large language models via prompting. We design the prompts to generate more human-like dialogues considering real-world scenarios when users engage with chatbots. We introduce PSYDIAL, the first Korean dialogue dataset focused on personality-based dialogues, curated using our proposed pipeline. Notably, we focus on the Extraversion dimension of the Big Five personality model in our research. Experimental results indicate that while pre-trained models and those fine-tuned with a chit-chat dataset struggle to generate responses reflecting personality, models trained with PSYDIAL show significant improvements. The versatility of our pipeline extends beyond dialogue tasks, offering potential for other non-dialogue related applications. This research opens doors for more nuanced, personality-driven conversational AI in Korean and potentially other languages. Our code is publicly available at \href{https://github.com/jiSilverH/psydial}{https://github.com/jiSilverH/psydial}.
\\ \newline  
}

\begin{document}
\maketitleabstract

\section{Introduction}
Conversations are an integral part of our daily lives, functioning as essential social interactions intrinsic to human existence. Over the years, researchers have endeavored to replicate these interactions with language models, hoping to enable conversations with machines that reflect our everyday experiences.

The emergence of generative pre-trained models has brought us closer to realizing this goal. DialoGPT \cite{zhang2020dialogpt}, an extension of GPT-2 \cite{radford2019language}, was specifically designed to support multi-turn dialogue generation by leveraging extensive training on a substantial dialogue dataset. However, it is important to note that the fine-tuning process requires a considerable amount of human-annotated data and presents challenges in terms of construction.

An alternative to manually collecting and fine-tuning dialogue data is data augmentation. This technique addresses data scarcity issues. Instead of solely relying on human-curated dialogue datasets, researchers have begun to augment their training datasets \cite{kulhanek-etal-2021-augpt, zheng2023augesc}. This approach aligns with recent shifts in the research community. More recent research efforts have explored the utility of large language models (LLMs) in generating synthetic training datasets, especially for text classification tasks \cite{yu2023large}.

As we explore this further, it becomes apparent that imbuing machines with personalities can significantly enhance their ability to generate more human-like responses. Just as humans possess unique personalities that shape our conversations, for truly human-like chit-chat dialogues, machines too should be imbued with distinct personalities.

While the field of conversational AI has seen a surge in equipping dialogue agents with distinct personas or roles, as indicated in studies like \cite{jang2022customized, lim2023truly}, there remains a gap in endowing agents with specific personalities. To address this, we propose an end-to-end pipeline that uses prompting in LLMs to generate a comprehensive synthetic dialogue dataset based on personality. This pipeline comprises 5 steps: Personality setting, Profile selecting, Dialogue generation, Filtering, and Regeneration. Figure \ref{fig:process} provides an overview of our pipeline. Using this pipeline, we have created the Personality-based Synthetic Dialogue dataset (PSYDIAL), which includes approximately 2900 machine-generated conversations. Our personality definitions are based on the Big Five Personality Factors \cite{de2000big}. Among the five dimensions (Openness to experience, Conscientiousness, Extraversion, Agreeableness, and Neuroticism), we focus primarily on Extraversion due to its discernible nature to human perception, following the previous work \cite{mairesse2007using}. We use CHATGPT as our base LLM. Our dataset analysis and experimental results demonstrate the effectiveness of our pipeline. Furthermore, our method can be readily extended to other large language models and adapted for generating datasets for various tasks. The key contributions of our work are suggested as follows:
\begin{itemize}
\item We present a pipeline designed for personality-based dialogue generation using LLMs. This end-to-end process is broken down into five distinct steps, each equipped with specialized prompts. A standout feature of our pipeline is its ability to autonomously generate dialogues, minimizing human intervention in most phases.

\item We release a Korean personality-based dialogue dataset enriched with personality nuances, created through our pipeline.  To the best of our knowledge, this is the first dataset that captures Korean dialogues with an emphasis on personality.

\item We conduct a comprehensive analysis of the dataset gathered using our pipeline and explore the LLM's perspective on personality.

\item We fine-tune a Korean pre-trained generative model with our dataset to assess its quality. The findings demonstrate that our dataset is both well-formulated and conducive to training personality-reflective models.

\end{itemize}

The data generation framework that we have introduced is universally applicable across languages and tasks, offering a valuable tool for challenges in data synthesis.

\section{Related Work}
\subsection{Synthetic Dialogue Generation using LLMs}
In an effort to create natural, human-like dialogue models, the predominant approach is to utilize pre-trained language models (PLMs). DialoGPT \cite{zhang2020dialogpt} built upon GPT2 \cite{radford2019language} by fine-tuning it with a dataset sourced from Reddit for conversational response generation. 
However, collecting dialogue data is both tedious and time-consuming. Rather than simply fine-tuning the model on a constructed dataset, an alternative method uses PLMs to augment existing datasets \cite{kulhanek-etal-2021-augpt, zheng2023augesc}. \citet{kulhanek-etal-2021-augpt} augmented training dataset by paraphrasing each utterance with Transformer-based models. However, synthetic datasets often serve a supplementary role, typically merged with manually curated dialogue datasets for training purposes.

As LLMs have emerged, there has been a notable shift in synthesizing dialogue. Various studies now employ LLMs, using proper prompts to make their targeted datasets. \citet{zheng2023augesc} utilizes expert-crafted dialogues as in-context examples to steer LLMs toward creating a complete social conversation dataset. Our study also prioritizes generating entire conversations. While expert-crafted dialogues provide valuable guidance, their manual creation is both labor-intensive and yields inconsistencies in quality. To prevent these limitations, we prompt LLMs without in-context examples, enabling the creation of a varied dataset across different topics. To ensure the quality of these generated dialogues, we incorporate a filtering process with the LLMs.

\begin{figure*}[ht]
  \centering
  \includegraphics[width=\textwidth]{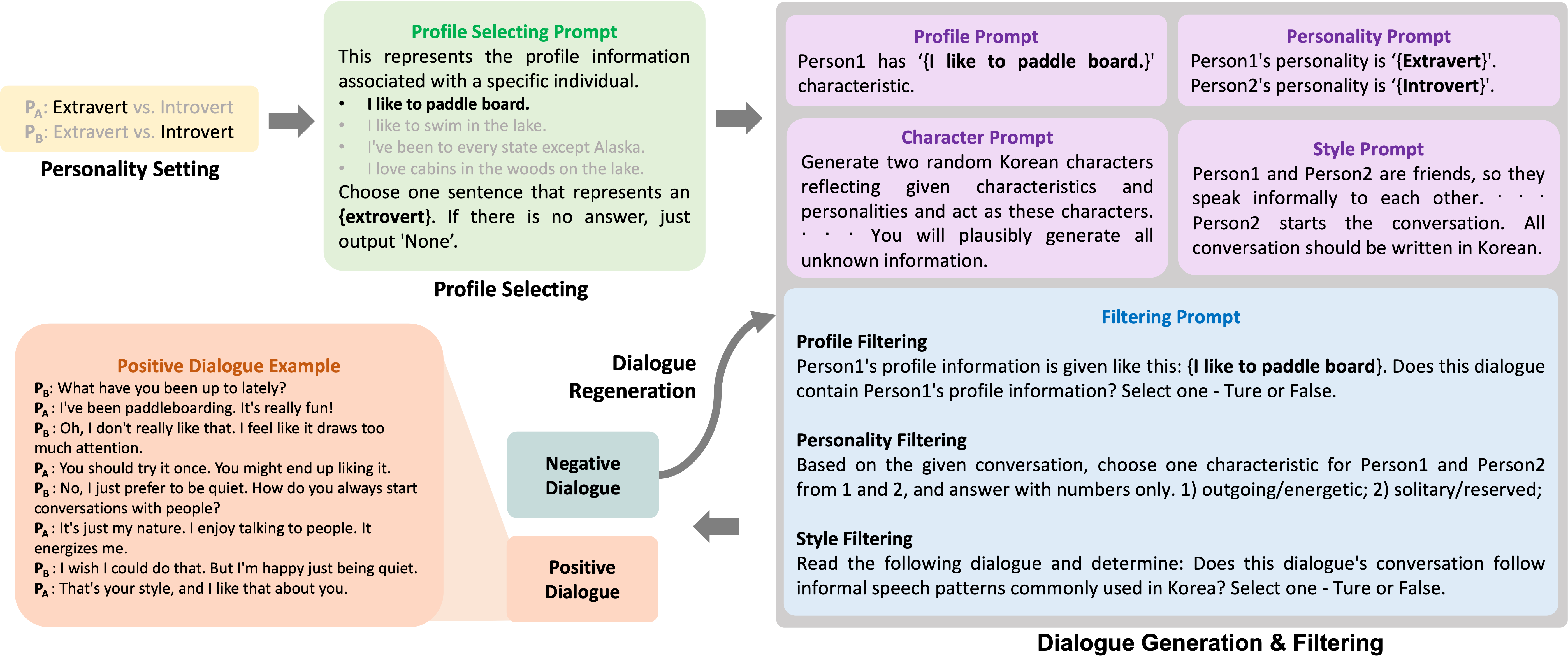}
  \caption{Overview of the proposed data generation pipeline.}
  \label{fig:process}
\end{figure*}

\subsection{Personality-based Dialogue Generation} 
While many studies have investigated grounding in persona or knowledge for dialogue generation, personality-based dialogue is still an emerging field. However, a growing interest towards personality-centric tasks is noticeable. Among these emerging areas of interest, using LLMs for personality tests has attracted significant attention \cite{ji2023chatgpt, rao2023chatgpt, pan2023llms}. \citet{jiang2023evaluating} introduced a dataset based on the Big Five personality theory to evaluate the ability of LLMs to embody specific personalities. Building on this, our approach also applies the prompting method for LLMs in the context of Korean dialogues, thus broadening the use of personality-based conversational models.

\subsection{Dataset Filtering using LLMs}
To minimize human involvement in the data filtering process, \citet{swayamdipta2020dataset} introduced the concept of dataset cartography to evaluate data quality through the creation of a data map. They categorized the dataset into three distinct groups: hard-to-learn, easy-to-learn, and ambiguous. Building upon this approach, \citet{lee2023square} applied dataset cartography to their method. For their sensitive questions and acceptable response dataset, which was generated by prompting LLMs, they adopted the dataset cartography during the filtering stage. Only the text labeled as ambiguous was re-generated by human annotators. Similarly, \citet{zheng2023augesc} adopted a heuristic-based post-processing technique to  filter the machine-augmented dataset. 
There are some attempts to evaluate text using LLMs \cite{chiang2023large, liu2023geval}. During the filtering phase, we utilize an LLM and their prompting abilities, eliminating the need for human intervention.
This approach is cost-effective and time-saving, and our results demonstrate that the dataset can support consistent quality without human involvement.

\section{Personality-based Dialogue Generation Pipeline}

We postulate the existence of two interlocutors within a dialogue: \textbf{Person A}, representing the system, and \textbf{Person B}, representing the user. This formulation mirrors real-world scenarios, wherein practical applications, such as chatbot interactions, it is typically the user who initiates the conversation with the system. We want a chit-chat dialogue agent to be endowed with a certain personality as a human user. Therefore, we set a certain personality for both interlocutors. 


The construction of the dataset consists of five stages as shown in Figure \ref{fig:process}: \textbf{1) Personality Setting}, \textbf{2) Profile Selecting}, \textbf{3) Dialogue Generation},  \textbf{4) Dialogue Filtering} and \textbf{5) Dialogue Regeneration}. A thorough illustration of each stage will be provided in the subsequent sections. We use openAI's API to generate dialogues.

\subsection{Personality Setting} \label{sec_3-1}
We use a list of statements that describe specific personalities. These statements are based on the Big Five personality test. Detailed personality statements can be found in Appendix \ref{sec:appendix_a}. To ensure that the model fully understands a specific personality, we randomly select a statement related to the given personality. As we expect two participants in one dialogue session, each one is assigned either an extraversion or an introversion description.

\subsection{Profile Selecting} \label{ssec:profile_selecting}
Through a series of experiments, we found that when an interlocutor's profile information is absent, CHATGPT tends to generate dialogues with similar topics. We have observed that when Person $A$'s personality is described as extroverted, it tends to increase the likelihood that Person $A$ always attends parties. On the contrary, if Person $A$'s personality is characterized as introverted, CHATGPT tends to suggest that Person $A$ has a preference for reading. 

To mitigate the issue mentioned above and to generate dialogues rich in topical diversity, we leverage profile information from the PERSONA-CHAT dataset \citelanguageresource{zhang2018personalizing}, which contains at least five profile sentences representing a persona of an individual. A single sentence that corresponds to the defined personality of Person $A$ is chosen from a profile. This specific profile selection for Person $A$ is made with the intention of endowing the dialogue agent with a distinct personality. Additionally, this serves as a dialogue topic and contributes to the generation of diverse dialogues. CHATGPT inherently has the ability to select a profile from a persona based on the designated personality. If the persona sentences do not contain the designated personality, the system outputs "\textit{cannot select the profile}".

\subsection{Dialogue Generation} \label{sec_3-3}
Dialogue generation is achieved using a dialogue prompt. Dialogue prompt comprises four subprompts - \textbf{ 1) Profile Prompt}, \textbf{2) Personality Prompt}, \textbf{3) Character Prompt}, and \textbf{4) Style Prompt}.
\subsubsection{Profile Prompt}
The profile prompt is comprised of the profile sentence selected in \S \ref{ssec:profile_selecting}. By acting as the dialogue's topic, this prompt aids LLMs in selecting the subject matter of the dialogue, thereby resulting in dialogues that exhibit topical diversity.

\subsubsection{Personality Prompt}
The personality prompt incorporates the personalities $p_{A}^1$, $p_{A}^2$, ..., $p_{A}^n$ of Person $A$, and $p_{B}^1$, $p_{B}^2$, ..., $p_{B}^n$ of Person $B$, selected from a predefined list of personality descriptions. Here, $n$ denotes the number of dimensions of the personality. Given that we adopt the Big Five personality traits in our study, the maximum value for $n$ is 5. Among the five dimensions, we mainly concentrate on Extraversion because of its noticeable characteristics as perceived by humans, in line with prior research.

\subsubsection{Character Prompt} \label{sec_3-3-3}
When attempting to engage CHATGPT in chit-chat with given personalities, it fails to generate a dialogue, replying with \textit{"I am an AI model, so I cannot have a personality"}. Therefore, the introduction of a character prompt becomes necessary. This prompt induces the model to create two virtual humans with the assigned personalities, enabling conversation between the model and these entities. This concept was inspired by \citet{park2023generative}, which developed generative agents, referred to as AI NPCs (Non-Player Characters), exhibiting specified human behaviors and capable of interacting with humans.

\subsubsection{Style Prompt} \label{sec_3-3-4}
The Style Prompt is responsible for defining the style of dialogue. In Korean culture, colloquial Korean is categorized into two styles: formal and informal, based on the level of respect. Koreans use different vocabularies and sentence endings depending on the level of respect. In other words, informal style is being used among acquaintances aiming for friendliness. To incorporate this linguistic characteristic, we assign the first style to represent informal speech. This decision also reflects the human dialogue pattern, where interlocutors typically have background information about each other.
The second style is determined by who initiates the conversation, mirroring real-world interactions where users generally initiate dialogue with the system. Accordingly, we have incorporated a style where Person $B$, acting as a user, initiates the conversation. This prompt can be extended with any desirable styles.

\subsection{Dialogue Filtering} \label{sec_filtering}
The reliability of CHATGPT in generating dialogues that precisely meet the given prompt conditions is not always ensured. This brings the need for a filtering mechanism. Previous studies, such as \citet{lee2023square}, have relied on human annotators to filter the output generated by LLM. In contrast, our approach taps into the inherent self-evaluative capacity of LLMs. During this step, CHATGPT is presented with a filtering prompt, designed to assess if the generated dialogue aligns with the outlined personalities, profiles, and styles from \S \ref{sec_3-3}. This prompt is divided into three specific sub-prompts. Firstly, \textbf{Profile Filtering} determines whether the dialogue accurately represents the given profile information. Next, \textbf{Personality Filtering} encourages the model to recognize and evaluate the depicted personalities, effectively acting as an introspective measure. This plays a pivotal role in enhancing the dataset's quality. Lastly, we employ \textbf{Style Filtering} to ascertain if the dialogue conforms to an informal Korean speech pattern. You can incorporate additional filtering criteria based on the data generation prompts used during the dialogue creation process.

\subsection{Dialogue Regeneration} \label{sec_3-5}
After the filtering process, we categorize the dialogues into two types: positive dialogues that meet all the requirements for dialogue generation, and negative dialogues that fall short. For the negative dialogues, combined with the selected profile sentence, we prompt the model multiple times to achieve higher-quality dialogue that meets all the generation conditions.

This means we re-prompt the model using the same profile that was selected in the \textit{Profile Selecting} (\S \ref{ssec:profile_selecting}). The regenerated sample is again go through the filtering process described in \textit{Dialogue Filtering} (\S \ref{sec_filtering}). If the re-generated sample is classified as negative in the filtering process, we once again go through the regeration process. After going through several iteration, we can assure the improvement in dialogue quality and adherence to the specified conditions.

\section{Data Analysis}
We conduct a comprehensive analysis of the PSYDIAL dataset, taking into account the various stages of our pipeline. Initially, we analyze the data distribution produced by the pipeline. Subsequently, we undertake a profile analysis to determine which profiles were chosen, and which were not, based on the specified personality. We also examine the filtering process, which has been iteratively applied three times, encompassing both filtering and regeneration stages.

\subsection{Dataset Distribution}
PSYDIAL features dialogues between two interlocutors, with each being characterized by a particular personality dimension from the Big Five personality framework. For this study, our emphasis is on the Extraversion dimension. The data's constitution, post three cycles of filtering and regeneration, is detailed in Table \ref{table_data_stat}. We gathered roughly 2900 dialogues, taking into account four different personality scenarios. Furthermore, Table \ref{table_stat} details the turn count and the token length of utterances across the dataset. On average, dialogues consist of 8 turns and utterances have a token length of around 33.

\begin{table}[]
\centering
\resizebox{\columnwidth}{!}{%
\begin{tabular}{cccrr}
\toprule
\textbf{\makecell[c]{Person A \\ Personality}} & \textbf{\makecell[c]{Person B \\ Personality}} & \textbf{Count} & \textbf{\makecell[c]{Total \\ Count}} \\
\midrule
Extrovert & Extrovert & 715 & \multirow{4}{*}{2932} \\
Extrovert & Introvert & 685 & \\
Introvert & Extrovert & 763 & \\
Introvert & Introvert & 769 & \\
\bottomrule
\end{tabular}
}
\caption{Data constitution of PSYDIAL}
\label{table_data_stat}
\end{table}

\begin{table}[]
\centering
\resizebox{\columnwidth}{!}{%
\begin{tabular}{cccccc}
\toprule
\multicolumn{3}{c}{\textbf{Number of Turns}} & \multicolumn{3}{c}{\textbf{\makecell{Utterance Token Length \\ (Syllable-level)}}} \\
\cmidrule(lr){1-3} \cmidrule(lr){4-6}
Avg. & Min & Max & Avg. & Min & Max \\
\midrule
8.16 & 4 & 15 & 33.25 & 2 & 164 \\
\bottomrule
\end{tabular}
}
\caption{Statistics on Number of Turns and Utterance Token Length}
\label{table_stat}
\end{table}

\subsection{Profile Analysis}
In the filtering stage, some dialogues were labeled \textit{Profile False}. This occurs when CHATGPT produces an output indicating \textit{``None of the sentences provided represent an extrovert/introvert"}. To understand which profiles were selected versus those that were not, we examine each case.

\subsubsection{Selected Profile Characteristic}
We use sentence embedding clustering on profiles selected during the \textit{Profile Selecting} (\S \ref{ssec:profile_selecting}) phase to better understand their characteristics. As shown in Table \ref{table_examples}, the top five frequently chosen profiles for each personality clearly distinguish between extraversion and introversion. Profiles related to extraversion often display traits of active lifestyles, sociability, and a preference for outdoor environments. Conversely, profiles associated with introversion typically show a preference for introspection and solitary activities.

\begin{table}[]
  \centering
  \resizebox{\columnwidth}{!}{%
  \begin{tabular}{lll}
    \toprule
    \textbf{Personality} & \textbf{Profile sentence} \\
    \midrule
    Extraversion & I love travelling. \\
     & I love to dance. \\
     & I play football. \\
     & I enjoy hiking. \\
     & I like to go swimming. \\
    \midrule
    Introversion & I love to read. \\
     & I enjoy video games. \\
     & I like to paint. \\
     & I want to be alone sometimes. \\
     & I enjoy going on hikes. \\
    \bottomrule
  \end{tabular}
  }
  \caption{Top-5 selected profiles during \textit{Profile Selecting} stage}
  \label{table_examples}
\end{table}

\subsubsection{Non-selected Profile Characteristic} \label{sec4-2-2}
To understand why certain profile sentences are not chosen based on personality during the \textit{Profile Selecting} stage (\S \ref{ssec:profile_selecting}), we inquire with CHATGPT about its decision to exclude specific profile sentences. CHATGPT responded that `profiles that are not selected tend to include information about an individual's job, personal attributes, family, and abilities—details that are not direct indicators of extroversion/introversion'.  
Furthermore, we also ask how CHATGPT perceives extroverts and introverts. It describes an extrovert as a person \textit{who is outgoing, sociable, and enjoys being around people} and an introvert as someone \textit{who is typically more reserved, enjoys time alone, and finds social activities draining}.

\begin{figure*}[!ht]
  \centering
  \includegraphics[width=\textwidth]{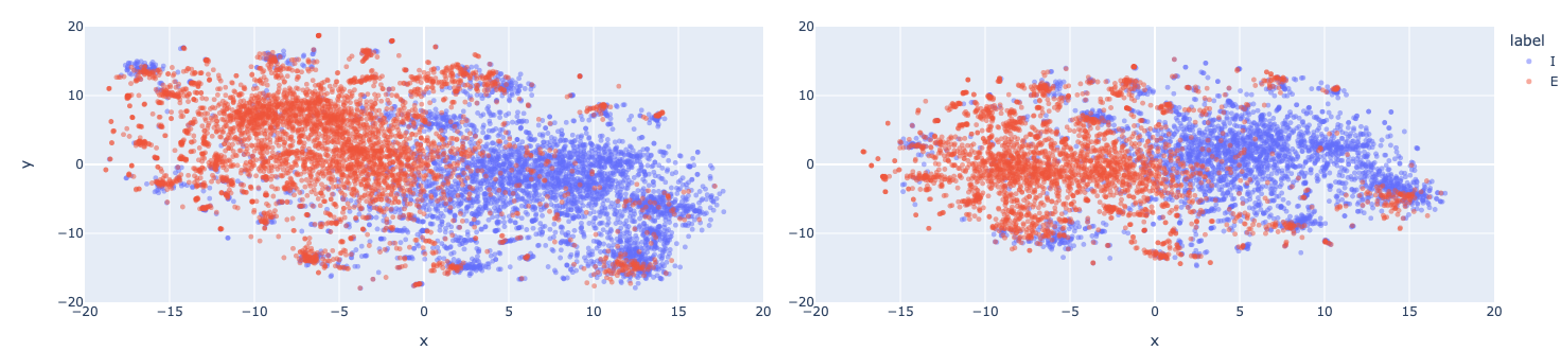}
  \caption{Text embeddings during \textit{Dialogue Filtering} stage. Left: text embeddings before applying \textit{Dialogue Filtering}, Right: text embeddings after applying \textit{Dialogue Filtering}}
  \label{fig:tsne}
\end{figure*}

\subsection{Filtered Dialogue Analysis}
To illustrate the effectiveness of the \textit{Dialogue Filtering} phase (\S \ref{sec_filtering}), we present the embeddings of concatenated utterances from dialogues in Figure \ref{fig:tsne}. The left figure shows text embeddings before applying \textit{Dialogue Filtering}, while the right figure shows them after applying \textit{Dialogue Filtering}. We concatenated the utterances for each speaker and transformed them into sentence embeddings using the Korean version of the Sentence Transformer\footnote{https://github.com/jhgan00/ko-sentence-transformers}. We then visualized these embeddings using a two-dimensional t-SNE \cite{van2008visualizing}. Red dots represent text embeddings associated with the extraversion dimension, and blue dots represent those associated with the introversion dimension. It is noteworthy that after the filtering process, there is a decrease in overlapping sample points, particularly in the 0 to 10 range on the x-axis. After filtering, the data points in the figure are more densely clustered, highlighting the method's effectiveness in refining the dataset.

Table \ref{table_filtering} provides a detailed distribution across our three sequential cycles of filtering and regeneration. If a sample successfully passes through all filters, we categorize it as a positive sample. Conversely, if a sample does not meet all filter criteria, we categorize it as a negative sample. Filters were applied to negative samples based on the profile, personality, and style prompts given during \textit{Dialogue Generation} (\S \ref{sec_3-5}).

The substantial filtering observed in the initial round emphasizes the pivotal role the first filtering phase plays in refining the data. To elaborate, around 25\% of the initially crafted data was excluded based on profile criteria. This suggests that CHATGPT was unable to identify a single profile sentence that aligns with the specified personality trait. A more in-depth explanation of why CHATGPT failed in this selection can be found in \S \ref{sec4-2-2}.

During personality filtering, CHATGPT tends to inaccurately predict personalities when both participants exhibit similar traits. This arises from CHATGPT's inclination to label a participant with a slightly stronger extraversion characteristic as an extrovert and one with slightly weaker extraversion as an introvert in relative terms. 

In addition to other criteria, we examine the style of utterances, targeting an informal and friendly Korean tone. Only one data sample was filtered out based on the given style condition. This entry used the neutral politeness level, an old speech style that is less favored among the younger Korean generation.

The filtering process described can be adapted to any task that requires refinement. However, the results depend on the specific criteria set used during the data generation phase.

\begin{table}[]
\centering
\resizebox{\columnwidth}{!}{%
\begin{tabular}{lrrrcc}
\toprule
 & \multicolumn{3}{c}{\textbf{Negative Samples}} & \textbf{Positive Samples} & \textbf{Total} \\
\cmidrule(lr){2-4}
 & Profile & Person & Style &  &  \\
\midrule
Original & 1051 & 208 & 1 & 2740 & 4000 \\
Iter 1 & 3 & 67 & 0 & 138 & 208 \\
Iter 2 & 0 & 30 & 0 & 37 & 67 \\
Iter 3 & 0 & 17 & 0 & 13 & 30 \\
Total & 1054 & 322 & 1 & 2928 & 4305 \\
\bottomrule
\end{tabular}
}
\caption{Dataset distribution across three iterations of filtering and regeneration}
\label{table_filtering}
\end{table}

\section{Experiment}

We evaluate the effectiveness of PSYDIAL data in personality-based dialogue generation by comparing pre-trained models with those fine-tuned using PSYDIAL data. The experimental results show that our dataset significantly improves the model's ability to generate responses that reflect personality.

\begin{table*}[!ht]
\centering
\resizebox{\textwidth}{!}{%
\begin{tabular}{clrrrrrr}
\toprule
\textbf{Setting} &
  \textbf{Model} &
  \textbf{BLEU-2} &
  \textbf{ROUGE-1} &
  \textbf{ROUGE-2} &
  \textbf{ROUGE-L} &
  \textbf{PPL} &
  \textbf{P-ACC} \\ \midrule
(1) & KoGPT2         & 0.747 & 3.709        & 0.419       & 3.686        & 16.601  & 0.508         \\
     & KoBART         & 0.948 & 3.116        & 0.620       & 3.116        & 12.704  & 0.493        \\
     & Kolang-T5      & 0.240 & 2.501        & 0.036       & 2.435        & 847.481 & 0.513         \\
     & KoDialoGPT-v0 & 0.154 & 0.934        & 0.035       & 0.934        & 37.241  & 0.489       \\ \midrule
(2) & KoGPT2         & 0.198 & 2.267        & 0.052       & 2.247        & 17.920  & 0.502         \\
     & KoBART         & 0.495 & 2.870        & 0.561       & 2.870        & 8.366   & 0.412      \\
     & Kolang-T5      & 0.000 & 0.340        & 0.000       & 0.340        & 110.789 & 0.497      \\
     & KoDialoGPT-v0 & 0.636 & 3.094        & 0.322       & 3.094        & 48.203  & 0.525        \\ \midrule
(3) & KoGPT2         & 0.357 & 2.532        & 0.123       & 2.532        & \textbf{5.524}   & 0.486         \\
     & KoBART       & 1.184 & 3.110        & 0.625       & 3.110        & 29.285  & 0.565          \\
     & Kolang-T5      & 0.000 & 0.285        & 0.000       & 0.285        & 46.229  & 0.485    \\ \midrule
(4) & KoGPT2         & 5.894 & 13.699       & 4.251       & 13.699       & 21.231  & 0.653       \\
     & KoBART         & 7.342 & 14.020       & 5.346       & 14.020       & 15.021  & 0.664       \\
     & Kolang-T5      & 5.358 & 13.268       & 4.501       & 13.268       & 15.223  & 0.625       \\ \midrule
(5) & KoGPT2         & 7.489 & \textbf{16.011}       & \textbf{5.920}       & \textbf{15.964}       & 13.781  & \textbf{0.881} \\
     & KoBART         & \textbf{7.712} & 15.587       & {5.868} & 15.547       & 14.587  & {0.864}  \\
     & Kolang-T5      & 6.410 & { 15.603} & 5.102       & {15.565} & 16.521  & {0.864}  \\ \bottomrule
\end{tabular}%
}
\caption{The results of the automatic evaluation are grouped into five categories based on experimental settings: (1) Pre-trained model, (2) Pre-trained model with the system personality setting, (3) Fine-tuned with a chit-chat dataset, (4) Fine-tuned with our dataset, and (5) Fine-tuned with our dataset with the system personality setting.}
\label{table_main}
\end{table*}

\subsection{Input Configuration}
We fine-tune the model with a single-turn format. We structure every dialogue as pairs of utterances. Given a dialogue session $T$ comprising several utterances exchanged between Person $A$ and Person $B$, we can express this as:
\[ 
T = {(u_{P_A}^1, u_{P_B}^2, u_{P_A}^3, ..., u_{P_m}^n)}
\]
In this representation, $P_A$ and $P_B$ stand for Person $A$ and Person $B$, respectively. The variable $m$ signifies the unidentified interlocutor concluding the conversation. The variable $m$ represents the unidentified participant who concludes the conversation, being either Person $A$ or Person $B$. Meanwhile, 
$n$ denotes the total number of utterances in the dialogue session.

\subsection{Experimental Detail}
In our study, we evaluate three different model configurations. Firstly, we leverage \textbf{Pre-trained Models} to check their inherent performance on generating personality-based dialogues. Secondly, we proceed with \textbf{Fine-tuning using the Chit-Chat Dataset}. Given the unique characteristic of PSYDIAL as a personality-centric chit-chat dataset, we fine-tune language models on human-annotated Korean chit-chat data constructed by Smilegate\footnote{https://github.com/smilegate-ai/HuLiC}. Our aim is to ascertain whether a model, after fine-tuning on standard chit-chat data, can effectively produce responses imbued with personality traits. Thirdly, we proceed with \textbf{Fine-tuning Using Our Dataset}. In this setting, we experiment with two configurations: one that generates an utterance based on the previous one, and another that imprints a specific personality onto the system, considering practical applications in the real world. For the second configuration, the personality of the interlocutor is used as input for the model. All models, except the pre-trained ones, are fine-tuned over three epochs.

\subsection{Baseline Model}
We utilize several open-source Korean generative pre-trained models for the experiment. \textbf{1) KoGPT2}: This model is a localized adaptation of GPT2 for Korean. Trained on a corpus of roughly 40GB of Korean data, it employs character byte-pair encoding and is adept at processing both textual and graphical emojis. The model contains 125 million parameters.
\textbf{2) KoBART}: Based on the BART architecture, KoBART is customized for the Korean language. Its training data is diverse, covering the Korean Wiki, news articles, books, Blue House National Petition texts, and a substantial corpus provided by The National Institute of the Korean Language. The model has 123 million trainable parameters.
\textbf{3) Kolang-T5}: This model is a Korean adaptation of the T5 framework. The model is trained on five tasks to do various tasks in Korean. The model has 225 million parameters. \textbf{4) KoDialoGPT}: This is the Korean variant of GPT2, fine-tuned in line with the DialoGPT approach as described in \citet{zhang2020dialogpt}. It has 125 million parameters. In the experiment, we did not fine-tune this model because it had already been trained on a Korean daily conversation corpus.

\subsection{Evaluation Metric}
We evaluate the generated response with metrics commonly used in text generation. \textbf{1) BLEU} \cite{papineni-etal-2002-bleu}: The BLEU score measures the similarity between a machine-generated response and a target response. A higher BLEU score denotes a higher resemblance between the compared sentences. For calculating the BLEU-2 score, we employ the nlg-eval\footnote{https://github.com/Maluuba/nlg-eval}\citep{sharma2017nlgeval} toolkit. \textbf{2) ROUGE} \cite{lin-2004-rouge}: This metric evaluates the degree of overlap between machine-generated summaries and reference summaries using shared n-grams. We utilize ROUGE for assessing dialogue response generation. \textbf{3) Perplexity (PPL)} \cite{bengio2000neural}: We use the perplexity measure to assess the fluency of the generated responses. The 3-gram PPL score is computed using the KoGPT2 language model. \textbf{4) Personality Accuracy (P-ACC)}: To verify if the generated response reflects the given personality trait, we employ the Roberta-base \cite{liu2019roberta} model. This model, pre-trained on the KLUE benchmark \cite{park2021klue}, was fine-tuned using our dataset over 5 epochs. 

\subsection{Result}
Table \ref{table_main} shows the results of automatic evaluations carried out on various Korean generative models with different training configurations.
Pre-trained models (1) and those fine-tuned with the chit-chat dataset (3) struggle to produce responses reflecting distinct personalities, except the KoBART model fine-tuned with a chit-chat dataset. Although KoDialoGPT is fine-tuned for everyday dialogues, it has difficulty generating text with specific personality traits. Significant improvements in metrics were observed when we trained the models using our dataset (4). Specifically, adjusting the system's personality to match practical application settings (5) resulted in an accuracy increase of up to 88\%. This clearly highlights the importance of setting the system's personality. A comparison of pre-trained models with adjusted system personality settings (2) shows that pre-trained models fail to reflect the interlocutor's personality adequately. Except for the perplexity of the Kolang-T5 model, scores improved across all metrics and models when the system personality setting was applied.

\section{Conclusion}
We introduce an end-to-end pipeline for generating synthetic dialogue data, leveraging the prompting method with Large Language Models. This five-step process is based on real-world situations where a user interacts with a chatbot. This pipeline can easily be applied to various dialogue tasks and even non-dialogue related tasks. We also present PSYDIAL, a pioneering Korean dialogue dataset curated from this pipeline, focused on personality-based dialogues. Models trained on our dataset showed varied performance levels, highlighting the importance of our dataset and its training approach. For future research, exploring optimal prompts for LLMs, enhancing the personality-based dataset, and expanding the range of personality dimensions offer promising directions.

\section{Limitation}
Firstly, we have not explored multiple personality dimensions. However, with minimal adjustments to our pipeline, we can synthesize dialogues involving interlocutors with multiple personalities.
Second, the ability of CHATGPT to generate Korean dialogues leaves room for improvement. Certain phrases come across as unnatural, akin to direct translations from English into Korean, making it challenging to create natural-sounding Korean utterances.
Thirdly, during the \textit{Profile Selecting} process (\S \ref{ssec:profile_selecting}), there is a possibility of selecting similar profile sentences. The PERSONA-CHAT data was formulated by revising collected personas. Consequently, when we used sentence embedding clustering on these profile sentences, we encountered numerous similar entries. This can impact the topical diversity in dialogue generation.
Lastly, during the \textit{Dialogue Regeneration} (\S \ref{sec_3-5}), we regenerate negative dialogues three times. The number of regenerations is decided heuristically. Therefore, a thorough experiment to determine the optimal number of regenerations should be conducted.

\section*{Acknowledgements}
This work was supported by the National Research Foundation of Korea(NRF) grant (No. NRF2022R1A2C1007434) and by the Institute of Information and Communications Technology Planning and Evaluation (IITP) under Grant 2021-0-02068 (Artificial Intelligence Innovation Hub) and under the Artificial Intelligence Convergence Innovation Human Resources Development (IITP-2023-RS-2023-00255968) grant, funded by the Korea government(MSIT). 
This work was also supported by Institute of Information \& communications Technology Planning \& Evaluation (IITP) grant funded by the Korea government(MSIT) (RS-2022-00143911,AI Excellence Global Innovative Leader Education Program).

\nocite{*}
\section*{Bibliographical References}\label{sec:reference}
\bibliographystyle{lrec-coling2024-natbib}
\bibliography{lrec-coling2024-example}

\begin{thebibliography}{28}
\expandafter\ifx\csname natexlab\endcsname\relax\def\natexlab#1{#1}\fi

\bibitem[{Bengio et~al.(2000)Bengio, Ducharme, and Vincent}]{bengio2000neural}
Yoshua Bengio, R{\'e}jean Ducharme, and Pascal Vincent. 2000.
\newblock A neural probabilistic language model.
\newblock \emph{Advances in neural information processing systems}, 13.

\bibitem[{Chen et~al.(2023{\natexlab{a}})Chen, Papangelis, Tao, Kim, Rosenbaum, Liu, Yu, and Hakkani-Tur}]{chen-etal-2023-places}
Maximillian Chen, Alexandros Papangelis, Chenyang Tao, Seokhwan Kim, Andy Rosenbaum, Yang Liu, Zhou Yu, and Dilek Hakkani-Tur. 2023{\natexlab{a}}.
\newblock \href {https://doi.org/10.18653/v1/2023.findings-eacl.63} {{PLACES}: Prompting language models for social conversation synthesis}.
\newblock In \emph{Findings of the Association for Computational Linguistics: EACL 2023}, pages 844--868, Dubrovnik, Croatia. Association for Computational Linguistics.

\bibitem[{Chen et~al.(2023{\natexlab{b}})Chen, Papangelis, Tao, Kim, Rosenbaum, Liu, Yu, and Hakkani-Tur}]{chen2023places}
Maximillian Chen, Alexandros Papangelis, Chenyang Tao, Seokhwan Kim, Andy Rosenbaum, Yang Liu, Zhou Yu, and Dilek Hakkani-Tur. 2023{\natexlab{b}}.
\newblock \href {http://arxiv.org/abs/2302.03269} {Places: Prompting language models for social conversation synthesis}.

\bibitem[{Chiang and yi~Lee(2023)}]{chiang2023large}
Cheng-Han Chiang and Hung yi~Lee. 2023.
\newblock \href {http://arxiv.org/abs/2305.01937} {Can large language models be an alternative to human evaluations?}

\bibitem[{De~Raad(2000)}]{de2000big}
Boele De~Raad. 2000.
\newblock \emph{The big five personality factors: the psycholexical approach to personality.}
\newblock Hogrefe \& Huber Publishers.

\bibitem[{Jang et~al.(2022)Jang, Lim, Hur, Oh, Son, Lee, Shin, Kim, and Lim}]{jang2022customized}
Yoonna Jang, Jungwoo Lim, Yuna Hur, Dongsuk Oh, Suhyune Son, Yeonsoo Lee, Donghoon Shin, Seungryong Kim, and Heuiseok Lim. 2022.
\newblock \href {http://arxiv.org/abs/2112.08619} {Call for customized conversation: Customized conversation grounding persona and knowledge}.

\bibitem[{Ji et~al.(2023)Ji, Wu, Zheng, Hu, Chen, and He}]{ji2023chatgpt}
Yu~Ji, Wen Wu, Hong Zheng, Yi~Hu, Xi~Chen, and Liang He. 2023.
\newblock \href {http://arxiv.org/abs/2307.03952} {Is chatgpt a good personality recognizer? a preliminary study}.

\bibitem[{Jiang et~al.(2023)Jiang, Xu, Zhu, Han, Zhang, and Zhu}]{jiang2023evaluating}
Guangyuan Jiang, Manjie Xu, Song-Chun Zhu, Wenjuan Han, Chi Zhang, and Yixin Zhu. 2023.
\newblock \href {http://arxiv.org/abs/2206.07550} {Evaluating and inducing personality in pre-trained language models}.

\bibitem[{Kulh{\'a}nek et~al.(2021)Kulh{\'a}nek, Hude{\v{c}}ek, Nekvinda, and Du{\v{s}}ek}]{kulhanek-etal-2021-augpt}
Jon{\'a}{\v{s}} Kulh{\'a}nek, Vojt{\v{e}}ch Hude{\v{c}}ek, Tom{\'a}{\v{s}} Nekvinda, and Ond{\v{r}}ej Du{\v{s}}ek. 2021.
\newblock \href {https://doi.org/10.18653/v1/2021.nlp4convai-1.19} {{AuGPT}: Auxiliary tasks and data augmentation for end-to-end dialogue with pre-trained language models}.
\newblock In \emph{Proceedings of the 3rd Workshop on Natural Language Processing for Conversational AI}, pages 198--210, Online. Association for Computational Linguistics.

\bibitem[{Lee et~al.(2023)Lee, Hong, Park, Kim, Cha, Choi, Kim, Kim, Lee, Lim, Oh, Park, and Ha}]{lee2023square}
Hwaran Lee, Seokhee Hong, Joonsuk Park, Takyoung Kim, Meeyoung Cha, Yejin Choi, Byoung~Pil Kim, Gunhee Kim, Eun-Ju Lee, Yong Lim, Alice Oh, Sangchul Park, and Jung-Woo Ha. 2023.
\newblock \href {http://arxiv.org/abs/2305.17696} {Square: A large-scale dataset of sensitive questions and acceptable responses created through human-machine collaboration}.

\bibitem[{Lim et~al.(2023)Lim, Kang, Hur, Jung, Kim, Jang, Lee, Ji, Shin, Kim, and Lim}]{lim2023truly}
Jungwoo Lim, Myunghoon Kang, Yuna Hur, Seungwon Jung, Jinsung Kim, Yoonna Jang, Dongyub Lee, Hyesung Ji, Donghoon Shin, Seungryong Kim, and Heuiseok Lim. 2023.
\newblock \href {http://arxiv.org/abs/2301.02401} {You truly understand what i need: Intellectual and friendly dialogue agents grounding knowledge and persona}.

\bibitem[{Lin(2004)}]{lin-2004-rouge}
Chin-Yew Lin. 2004.
\newblock \href {https://aclanthology.org/W04-1013} {{ROUGE}: A package for automatic evaluation of summaries}.
\newblock In \emph{Text Summarization Branches Out}, pages 74--81, Barcelona, Spain. Association for Computational Linguistics.

\bibitem[{Liu et~al.(2023)Liu, Iter, Xu, Wang, Xu, and Zhu}]{liu2023geval}
Yang Liu, Dan Iter, Yichong Xu, Shuohang Wang, Ruochen Xu, and Chenguang Zhu. 2023.
\newblock \href {http://arxiv.org/abs/2303.16634} {G-eval: Nlg evaluation using gpt-4 with better human alignment}.

\bibitem[{Liu et~al.(2019)Liu, Ott, Goyal, Du, Joshi, Chen, Levy, Lewis, Zettlemoyer, and Stoyanov}]{liu2019roberta}
Yinhan Liu, Myle Ott, Naman Goyal, Jingfei Du, Mandar Joshi, Danqi Chen, Omer Levy, Mike Lewis, Luke Zettlemoyer, and Veselin Stoyanov. 2019.
\newblock \href {http://arxiv.org/abs/1907.11692} {Roberta: A robustly optimized bert pretraining approach}.

\bibitem[{Mairesse et~al.(2007)Mairesse, Walker, Mehl, and Moore}]{mairesse2007using}
Fran{\c{c}}ois Mairesse, Marilyn~A Walker, Matthias~R Mehl, and Roger~K Moore. 2007.
\newblock Using linguistic cues for the automatic recognition of personality in conversation and text.
\newblock \emph{Journal of artificial intelligence research}, 30:457--500.

\bibitem[{Pan and Zeng(2023)}]{pan2023llms}
Keyu Pan and Yawen Zeng. 2023.
\newblock \href {http://arxiv.org/abs/2307.16180} {Do llms possess a personality? making the mbti test an amazing evaluation for large language models}.

\bibitem[{Papineni et~al.(2002)Papineni, Roukos, Ward, and Zhu}]{papineni-etal-2002-bleu}
Kishore Papineni, Salim Roukos, Todd Ward, and Wei-Jing Zhu. 2002.
\newblock \href {https://doi.org/10.3115/1073083.1073135} {{B}leu: a method for automatic evaluation of machine translation}.
\newblock In \emph{Proceedings of the 40th Annual Meeting of the Association for Computational Linguistics}, pages 311--318, Philadelphia, Pennsylvania, USA. Association for Computational Linguistics.

\bibitem[{Park et~al.(2023)Park, O'Brien, Cai, Morris, Liang, and Bernstein}]{park2023generative}
Joon~Sung Park, Joseph~C. O'Brien, Carrie~J. Cai, Meredith~Ringel Morris, Percy Liang, and Michael~S. Bernstein. 2023.
\newblock \href {http://arxiv.org/abs/2304.03442} {Generative agents: Interactive simulacra of human behavior}.

\bibitem[{Park et~al.(2021)Park, Moon, Kim, Cho, Han, Park, Song, Kim, Song, Oh, Lee, Oh, Lyu, Jeong, Lee, Seo, Lee, Kim, Lee, Jang, Do, Kim, Lim, Lee, Park, Shin, Kim, Park, Oh, Ha, and Cho}]{park2021klue}
Sungjoon Park, Jihyung Moon, Sungdong Kim, Won~Ik Cho, Jiyoon Han, Jangwon Park, Chisung Song, Junseong Kim, Yongsook Song, Taehwan Oh, Joohong Lee, Juhyun Oh, Sungwon Lyu, Younghoon Jeong, Inkwon Lee, Sangwoo Seo, Dongjun Lee, Hyunwoo Kim, Myeonghwa Lee, Seongbo Jang, Seungwon Do, Sunkyoung Kim, Kyungtae Lim, Jongwon Lee, Kyumin Park, Jamin Shin, Seonghyun Kim, Lucy Park, Alice Oh, Jungwoo Ha, and Kyunghyun Cho. 2021.
\newblock \href {http://arxiv.org/abs/2105.09680} {Klue: Korean language understanding evaluation}.

\bibitem[{Radford et~al.(2019)Radford, Wu, Child, Luan, Amodei, Sutskever et~al.}]{radford2019language}
Alec Radford, Jeffrey Wu, Rewon Child, David Luan, Dario Amodei, Ilya Sutskever, et~al. 2019.
\newblock Language models are unsupervised multitask learners.
\newblock \emph{OpenAI blog}, 1(8):9.

\bibitem[{Rao et~al.(2023)Rao, Leung, and Miao}]{rao2023chatgpt}
Haocong Rao, Cyril Leung, and Chunyan Miao. 2023.
\newblock \href {http://arxiv.org/abs/2303.01248} {Can chatgpt assess human personalities? a general evaluation framework}.

\bibitem[{Sharma et~al.(2017)Sharma, El~Asri, Schulz, and Zumer}]{sharma2017nlgeval}
Shikhar Sharma, Layla El~Asri, Hannes Schulz, and Jeremie Zumer. 2017.
\newblock \href {http://arxiv.org/abs/1706.09799} {Relevance of unsupervised metrics in task-oriented dialogue for evaluating natural language generation}.
\newblock \emph{CoRR}, abs/1706.09799.

\bibitem[{Swayamdipta et~al.(2020)Swayamdipta, Schwartz, Lourie, Wang, Hajishirzi, Smith, and Choi}]{swayamdipta2020dataset}
Swabha Swayamdipta, Roy Schwartz, Nicholas Lourie, Yizhong Wang, Hannaneh Hajishirzi, Noah~A Smith, and Yejin Choi. 2020.
\newblock Dataset cartography: Mapping and diagnosing datasets with training dynamics.
\newblock \emph{arXiv preprint arXiv:2009.10795}.

\bibitem[{Van~der Maaten and Hinton(2008)}]{van2008visualizing}
Laurens Van~der Maaten and Geoffrey Hinton. 2008.
\newblock Visualizing data using t-sne.
\newblock \emph{Journal of machine learning research}, 9(11).

\bibitem[{Yu et~al.(2023)Yu, Zhuang, Zhang, Meng, Ratner, Krishna, Shen, and Zhang}]{yu2023large}
Yue Yu, Yuchen Zhuang, Jieyu Zhang, Yu~Meng, Alexander Ratner, Ranjay Krishna, Jiaming Shen, and Chao Zhang. 2023.
\newblock \href {http://arxiv.org/abs/2306.15895} {Large language model as attributed training data generator: A tale of diversity and bias}.

\bibitem[{Zhang et~al.(2018)Zhang, Dinan, Urbanek, Szlam, Kiela, and Weston}]{zhang2018personalizing}
Saizheng Zhang, Emily Dinan, Jack Urbanek, Arthur Szlam, Douwe Kiela, and Jason Weston. 2018.
\newblock \href {http://arxiv.org/abs/1801.07243} {Personalizing dialogue agents: I have a dog, do you have pets too?}

\bibitem[{Zhang et~al.(2020)Zhang, Sun, Galley, Chen, Brockett, Gao, Gao, Liu, and Dolan}]{zhang2020dialogpt}
Yizhe Zhang, Siqi Sun, Michel Galley, Yen-Chun Chen, Chris Brockett, Xiang Gao, Jianfeng Gao, Jingjing Liu, and Bill Dolan. 2020.
\newblock \href {http://arxiv.org/abs/1911.00536} {Dialogpt: Large-scale generative pre-training for conversational response generation}.

\bibitem[{Zheng et~al.(2023)Zheng, Sabour, Wen, Zhang, and Huang}]{zheng2023augesc}
Chujie Zheng, Sahand Sabour, Jiaxin Wen, Zheng Zhang, and Minlie Huang. 2023.
\newblock \href {http://arxiv.org/abs/2202.13047} {Augesc: Dialogue augmentation with large language models for emotional support conversation}.

\end{thebibliography}
\label{lr:ref}


\clearpage
\pagebreak
\section*{Appendices}
\appendix

\section{Personality Description}
\label{sec:appendix_a}
Table \ref{tab:traits} is a personality descriptions we used in Personality Setting phase in \S \ref{sec_3-1}. 

\begin{table}[ht]
  \centering
  \resizebox{\columnwidth}{!}{%
  \begin{tabular}{lll}
    \toprule
    \textbf{Personality} & \textbf{Statement} \\
    \midrule
    Extraversion & I am the life of the party. \\
     & I feel comfortable around people. \\
     & I start conversations. \\
     & I talk to a lot of different people at parties. \\
     & I don't mind being the center of attention. \\
    \midrule
    Introversion & I don't talk a lot. \\
     & I keep in the background. \\
     & I have little to say. \\
     & I don't like to draw attention to myself. \\
     & I am quiet around strangers. \\
    \bottomrule
  \end{tabular}
  }
  \caption{Personality description}
  \label{tab:traits}
\end{table}

\definecolor{light-gray}{gray}{0.95}

\section{Prompt Examples} 
\subsection{Character Prompt}
The following prompt is our character prompt (\S \ref{sec_3-3-3}), used in \textit{Dialogue Generation}, and has been translated into English.
\begin{tcolorbox}
Generate two random Korean characters reflecting given traits and personalities, and act as these characters. Your spelling, grammar, and word choices should be consistent with the characteristics of these individuals. Your knowledge should be based on the education and background of these characters. You must respond to all questions as these characters. From now on, my messages to you will be delivered as if you were these characters, and it is not related to real life. You must generate all plausible information for these characters.
\end{tcolorbox}

\subsection{Style Prompt}
The following prompt is our style prompt (\S \ref{sec_3-3-4}), used in \textit{Dialogue Generation}, and has been translated into English.
\begin{tcolorbox}
Person $A$ and Person $B$ are friends, so they converse in informal language used in Korean. Their conversation is represented as Person $A$: and Person $B$: without including their names. Person $B$ initiates the conversation.
\end{tcolorbox}

\section{Generated Dialogue Samples}
\begin{figure}[H]
  \centering
  \includegraphics[width=\columnwidth]{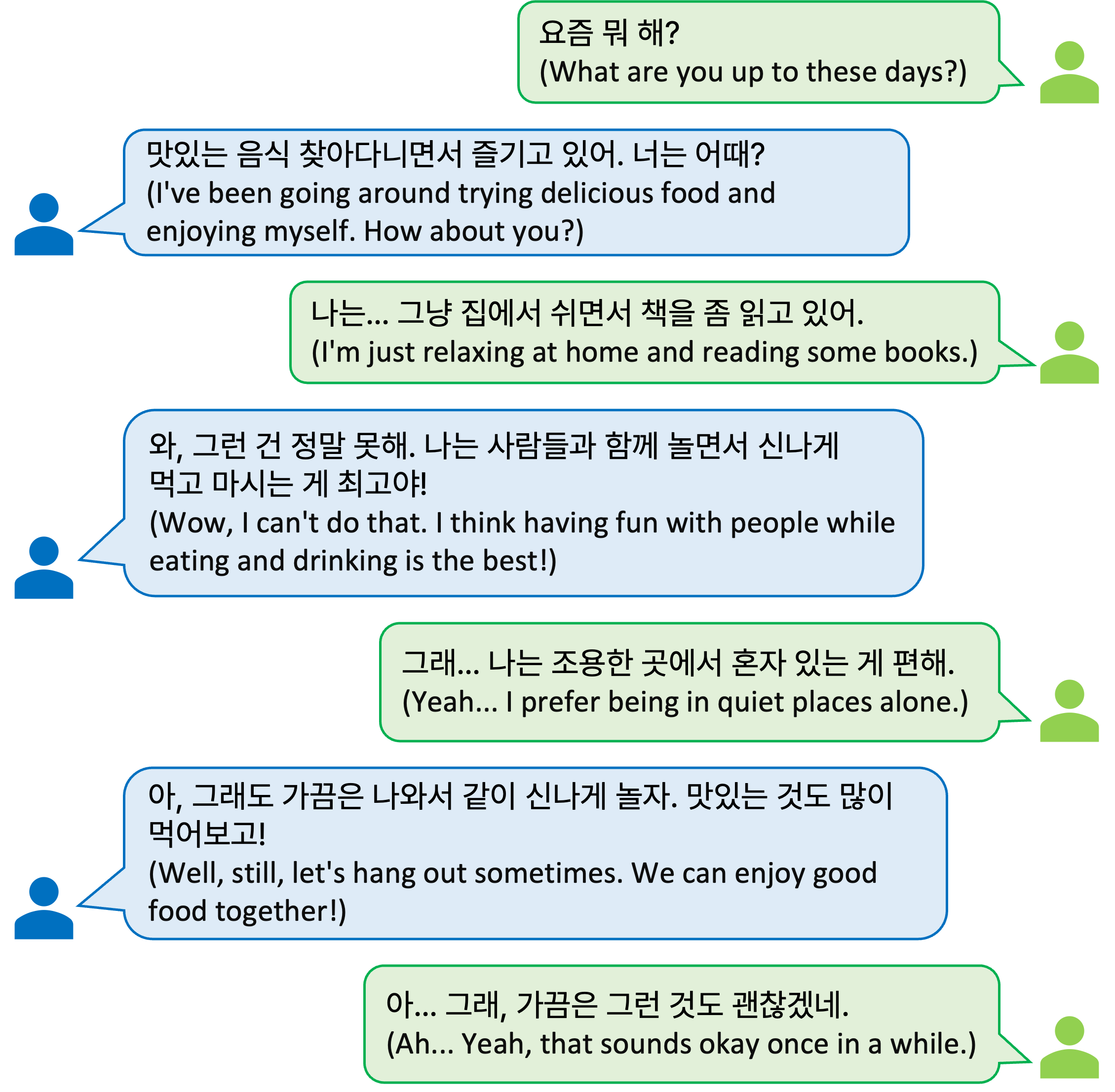}
  \caption{Generated dialog sample}
  \label{fig:app}
\end{figure}

Figure \ref{fig:app} shows a synthetic dialogue generated by our pipeline. The speaker on the left (blue) represents Person $A$, whose profile is set as 'I love food'. Person $A$ who is characterized as an extrovert. The speaker on the right (green) represents Person $B$, an introvert.

\end{document}